# Time series modeling with pruned multi-layer perceptron and 2-stage damped least-squares method

**Cyril Voyant[1,2], Wani Tamas[2], Christophe Paoli[2], Aurélia Balu[2], Marc Muselli[2], Marie-Laure Nivet[2], Gilles Notton[2]**

1-CHD Castelluccio, radiophysics unit, 20000 Ajaccio (France)

2-University of Corsica, *UMR* CNRS *6134 SPE*, Campus Grimaldi, BP 52, 20250 Corte, (France)

Cooresponding author, Cyril Voyant : voyant@univ-corse.fr

**Abstract**. A Multi-Layer Perceptron (MLP) defines a family of artificial neural networks often used in TS modeling and forecasting. Because of its "black box" aspect, many researchers refuse to use it. Moreover, the optimization (often based on the exhaustive approach where "all" configurations are tested) and learning phases of this artificial intelligence tool (often based on the Levenberg-Marquardt algorithm; LMA) are weaknesses of this approach (exhaustively and local minima). These two tasks must be repeated depending on the knowledge of each new problem studied, making the process, long, laborious and not systematically robust. In this paper a pruning process is proposed. This method allows, during the training phase, to carry out an inputs selecting method activating (or not) inter-nodes connections in order to verify if forecasting is improved. We propose to use iteratively the popular damped least-squares method to activate inputs and neurons. A first pass is applied to 10% of the learning sample to determine weights significantly different from 0 and delete other. Then a classical batch process based on LMA is used with the new MLP. The validation is done using 25 measured meteorological TS and cross-comparing the prediction results of the classical LMA and the 2-stage LMA.

## 1. Background

The primary goal of time series (TS) analysis is forecasting, i.e. using the past to predict the future [1]. This formalism is used in many scientific fields like econometrics, seismology or meteorology. Lot of methods are dedicated to the prediction of discrete phenomena, one of the most popular is the artificial neural network (ANN) [1,2]. From a mathematical point of view, ANN is a function defined as the composition of other functions [3]. Members of the class of such functions are obtained by varying parameters, (as connections or weights). A Multi-Layer Perceptron (MLP) defines a family of functions often used in TS modeling [1]. In this model, neurons are grouped in layers and only forward connections exist. A typical MLP consists of an input, hidden and output layers, including neurons, weights and a transfer functions. Each neuron (noted *i*) transforms the weighted sum (weight $w_{ij}$, bias $b_i$) of inputs ($x_j$) into an output ($y_i = f(\sum_{j=1}^{n} x_j w_{ij} + b_i)$) using a transfer function (*f*). The goal of this method is to determine weights and bias for a given problem. A complex process is necessary to adapt connections using a suitable training algorithm (often based on the Levenberg-Marquardt algorithm; LMA [4,5]). The training step is dependent of the number of inputs, layers and hidden nodes. The better configuration defines the optimized MLP [1]. This step is the weaknesses of this approach because no consensus or scientific rules exist, often the use of the exhaustive approach (where "all" configurations are tested) is the only usable and must be repeated for new studied problem, making the process, long, laborious and not systematically robust. This "black box" aspect leads any researchers to refuse to use it. In this paper a pruning process allowing to automatically selecting inputs is proposed. This method allows, during the training phase, to carry out a selecting method activating (or not) inter-nodes connections. With this process, the optimization step becomes self-acting and the parsimony principle is kept. Less the MLP is complex more it is efficient [3].

## 2. Materials and methods

We propose to use iteratively the popular LMA also known as the damped least-squares method to activate the inputs and neurons ($m$ weights and bias) [6]. A first pass is applied to 10% of the learning sample. For each step, a system of $m$ non-linear equations with $m$ unknowns is solved (see equation 1 in case of 1 hidden layer MLP where $O$ and $I$ are the outputs and inputs, $W^1$, $B^1$ and $W^2$, $B^2$ the weights and bias matrices of the hidden and output layer).

$$O^i = W^2.(\tanh(W^1.I^i + B^1)) + B^2 \text{ with } 1<i<m, \qquad (1)$$

After this first phase, each weights and bias ($\omega_{i \in [1,m]}$) are represented by probability distributions. A statistical test based on the bootstrap distribution is used to determine if the first moment of each $\omega_i$ is significantly different from 0 [7]. Before to initiate the second pass, the network is customized and connections related to each $\omega_i$ non-significantly different from 0 are canceled. Then a classical batch process based on LMA is used with the new MLP. We validated our method with 5 hourly meteorological TS (wind direction *WD*, wind speed *WS*, Global radiation *Glo*, Humidity *Hum* and temperature *Tem*), each one measured in 5 French sites (Ajaccio, Bastia, Corte, Marseille and Nice). Note that no pretreatment are operated and that different non-stationarities and periodicities are present. For all TS and locations, we used 3200 measures for the training and 400 for the cross comparison of the classical LMA and the 2-stage LMA during the year 2008.

### 2.1. First pass

The first stage begins with the generation of $N$ (10% of the total data used during training randomly chosen) systems of $m$ nonlinear equations with $m$ unknowns (MLP constructed with $m$ weights and bias). The method chosen for solve this problem is the LMA method and the ad-hoc objective function $F$ (mean square error between calculations and measures). It is an approximation of the Gauss-newton method, the result of the k$^{th}$ iterations ($\omega_k$) corresponding to the local minimum of the function $F$ is generated by the linear set of equations (1) [4–6,8]:

$$(J(\omega_k)^T.J(\omega_k) + \lambda_k I).d_k = -J(\omega_k)^T.F(\omega_k) \qquad (2)$$

$J$ denotes the Jacobian matrix of $F$ and the scalar $\lambda_k$ controls both the magnitude and variation $d_k$ ($d_k = \Delta\omega = \omega_{k+1} - \omega_k$). After the $N$ solving, all the weights and bias are represented by a distribution which will be used during the second stage of the second stage of the methodology. Note that these distributions are not all normal (according to the Jarque-Bera test).

### 2.2. Second pass

Before to initiate the second pass, the network is customized and connections related to each $\omega_i$ non-significantly different from 0 are canceled. We use confidence intervals from the bootstrap distributions (4000 samples) of the weights and bias parameters [7]. The rules of selection (directly linked to the $\alpha$ value of significance model) is based on the product of the two endpoints $t_1$ and $t_2$ defined respectively by the $\alpha/2^{th}$ and $(1-\alpha/2)^{th}$ percentiles of the distribution. If $t_1 \cdot t_2 < 0$ the weight (or bias) is considered as non-significantly different from 0, else it is considered different from 0. The connection of the MLP corresponding to the first case ($t_1 \cdot t_2 < 0$) are cancelled, other are kept. The pruned MLP (noted pMLP) is then trained with the classical LMA.

## 3. Results

In the Figure 1a is represented the box plot of the nRMSE distribution concerning the five meteorological parameters and the five studied cities. For each case, seven runs are operated, so 175 manipulations are performed with the pruned methodology describe above (at left in the Figure 1 and noted pMLP) and the standard approach (at right in the Figure 1 and noted MLP). The chosen architecture is the same for all cases: 7 inputs representing the seven first lags of the meteorological parameter tested and 2 hidden nodes (only one hidden layer). In this figure, we see that the first, the second and the third quartiles are equivalent; thereby we understand that a lot of connections and weights are superfluous. The results of all the simulations are represented in the Figure 1b. Only the points positioned in the upper zone are related to "pMLP is better than MLP" cases are plotted. It appears the points are closer to the y=1 curve in the top area rather than in the bottom area, but this observation seems insignificant.

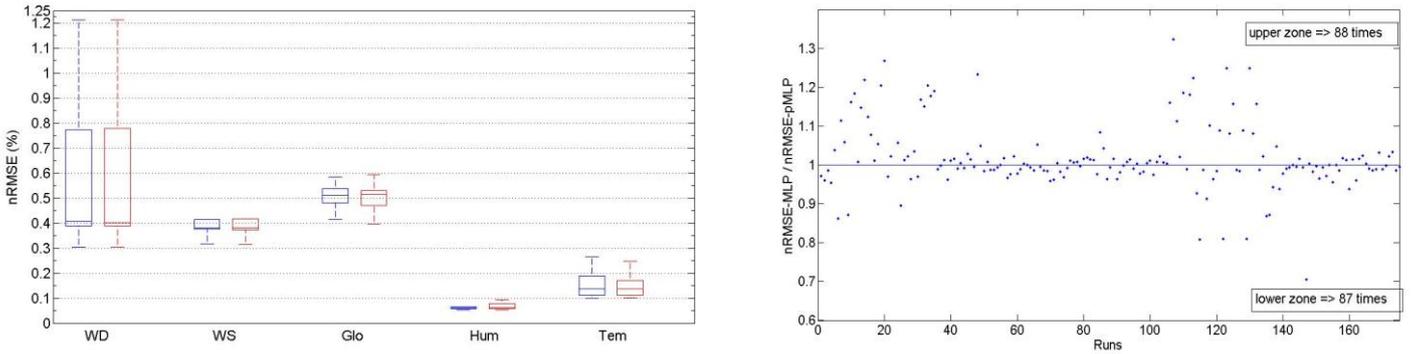

**Figure 1: a. nRMSE distribution comparison related to pMLP (at left) and MLP (at right).**
 **b. ratio of the nRMSE generated by MLP and pMLP**

The table 1 exposes the results for all locations and parameters of the MLP and pMLP approach, the minimum of the nRMSE and nMAE [1] through the seven runs.

| Data | City | MLP | | pMLP | | |
|---|---|---|---|---|---|---|
| | | *nRMSE* | *nMAE* | Ratio pruning | *nRMSE* | *nMAE* |
| WD | Aja | 0.765 | 0.463 | 0.27 | **0.762** | **0.461** |
| | Bas | 0.393 | 0.257 | 0.27 | **0.388** | **0.252** |
| | Cor | **1.191** | 0.941 | 0.21 | 1.194 | **0.931** |
| | Mar | **0.387** | **0.256** | 0.20 | 0.388 | 0.257 |
| | Nic | **0.304** | **0.199** | 0.20 | 0.305 | 0.211 |
| WS | Aja | **0.399** | **0.302** | 0.11 | 0.410 | 0.308 |
| | Bas | 0.374 | **0.277** | 0.18 | **0.370** | 0.279 |
| | Cor | **1.148** | 0.910 | 0.22 | 1.183 | **0.902** |
| | Mar | 0.377 | 0.264 | 0.21 | **0.377** | **0.263** |
| | Nic | 0.318 | 0.206 | 0.23 | **0.314** | **0.202** |
| Glo | Aja | 0.525 | 0.413 | 0.17 | **0.472** | **0.372** |
| | Bas | **0.439** | **0.323** | 0.21 | 0.456 | 0.370 |
| | Cor | **0.298** | **0.214** | 0.21 | 0.323 | 0.250 |
| | Mar | 0.416 | 0.346 | 0.21 | **0.378** | **0.302** |
| | Nic | **0.455** | **0.380** | 0.20 | 0.465 | 0.386 |
| Hum | *Aja* | **0.064** | 0.049 | 0.22 | **0.064** | **0.048** |
|     | *Bas* | **0.061** | **0.044** | 0.20 | **0.061** | 0.045 |
|     | *Cor* | **0.055** | **0.036** | 0.23 | 0.056 | 0.037 |
|     | *Mar* | **0.053** | **0.036** | 0.23 | **0.053** | **0.036** |
|     | *Nic* | **0.083** | **0.059** | 0.19 | **0.083** | **0.059** |
| Tem | *Aja* | **0.099** | **0.077** | 0.29 | 0.101 | **0.077** |
|     | *Bas* | 0.113 | 0.080 | 0.25 | **0.111** | **0.079** |
|     | *Cor* | 0.224 | 0.158 | 0.29 | **0.207** | **0.133** |
|     | *Mar* | 0.111 | 0.079 | 0.28 | **0.109** | **0.073** |
|     | *Nic* | 0.147 | **0.108** | 0.21 | 0.146 | 0.111 |

**Table 1: nRMSE and nMAE minima for all location and all parameters, in bold the better results between MLP and pMLP**

pMLP is very slightly better than MLP, in 60% of cases the nMRSE and the nMAE are the lowest. Note that the pruning concerns about 20% of weights and bias.

## 4. Conclusions

The 2-pass approach improves slightly the forecasting quality. In average, 20% of the connections are removed with this approach. According to the parsimony principle, these simplifications increase the generalization capacity and should allow building a robust predictor [2].

This first study done in a quasi-optimized case (7 inputs and 2 hidden neurons) precedes a more general one, where a standard MLP (more than 15 inputs and 15 hidden nodes) will be studied. Indeed, we have shown that the pruning method presented here is able to simplify the network while the performance is roughly equivalent. Applying the 2-pass approach in a 15x15 MLP should allow to optimize it without apply the exhaustive test where all the architectures are try out. Moreover, for users it is a totally transparent methodology, suitable for all TS and faster than classical optimization process. According to the conclusion of this study, it possible that results based on the 2-stage approach may be better than the classical approach based on the "1-stage" LM algorithm.